\documentclass[10pt,twocolumn,letterpaper]{article}

\usepackage[pagenumbers]{wacv}

\usepackage[accsupp]{axessibility}
\usepackage{graphicx}
\usepackage{booktabs}
\usepackage{array}
\usepackage{float}
\usepackage{dblfloatfix}

\definecolor{wacvblue}{rgb}{0.21,0.49,0.74}
\usepackage[breaklinks,colorlinks,allcolors=wacvblue]{hyperref}
\usepackage{placeins}
\usepackage[most]{tcolorbox}
\usepackage{listings}

\colorlet{promptwrap}{wacvblue!75!black}

\lstdefinestyle{jsonbox}{
    basicstyle=\scriptsize\ttfamily,
    breaklines=true,
    breakatwhitespace=false,
    breakindent=0pt,
    alsoother=_,
    postbreak=\mbox{\textcolor{promptwrap}{$\hookrightarrow$}\space},
    columns=fullflexible,
    keepspaces=true,
    showstringspaces=false,
    frame=none,
    xleftmargin=0pt,
    xrightmargin=0pt,
    aboveskip=0pt,
    belowskip=0pt}

\lstnewenvironment{promptjson}{\lstset{style=jsonbox}}{}

\newtcolorbox{promptbox}[1]{
    enhanced,
    colback=wacvblue!3,
    colframe=wacvblue!75!black,
    boxrule=0.55pt,
    arc=1.2mm,
    left=1.6mm,
    right=1.6mm,
    top=0.5mm,
    bottom=0.5mm,
    fonttitle=\bfseries\normalsize,
    fontupper=\footnotesize,
    title={#1}}

\newcolumntype{P}[1]{>{\raggedright\arraybackslash}p{#1}}

\def\wacvPaperID{864}
\def\confName{WACV}
\def\confYear{2027}

\title{RadOncGym: An Agent-Based Framework for Radiotherapy Dose-Mimicking Optimization}

\hypersetup{
    pdftitle={RadOncGym: An Agent-Based Framework for Radiotherapy Dose-Mimicking Optimization},
    pdfauthor={Tong Miao, Shupeng Cheng, Xiwen Chen, Ruishan Liu}
}

\author{%
Tong Miao\thanks{Equal contributions. $^{\dagger}$Corresponding author.}\quad
Shupeng Cheng\footnotemark[1]\quad
Xiwen Chen\footnotemark[1]\quad
Ruishan Liu$^{\dagger}$\\
{\normalsize University of Southern California, USA}\\
{\small\ttfamily \href{mailto:miaot@usc.edu}{\textcolor{black}{miaot@usc.edu}}, \href{mailto:shupengc@usc.edu}{\textcolor{black}{shupengc@usc.edu}}, \href{mailto:xiwenc@usc.edu}{\textcolor{black}{xiwenc@usc.edu}}, \href{mailto:ruishanl@usc.edu}{\textcolor{black}{ruishanl@usc.edu}}}%
}

\makeatletter
\newcommand*{\ps@arxivfirst}{%
    \ps@plain
    \def\@oddfoot{%
        \rlap{\raisebox{-18pt}[0pt][0pt]{%
            \parbox[t]{\textwidth}{\normalfont\fontsize{6.35pt}{8pt}\selectfont\raggedright\color{black!85}
            \textcopyright{} 2027 IEEE. Personal use of this material is permitted. Permission from IEEE must be obtained for all other uses, in any current or future media, including reprinting/republishing this material for advertising or promotional purposes, creating new collective works, for resale or redistribution to servers or lists, or reuse of any copyrighted component of this work in other works.
            \par}%
        }}%
        \normalfont\hfil\thepage\hfil
    }%
    \let\@evenfoot\@oddfoot
}
\makeatother

\begin{document}
\maketitle
\thispagestyle{arxivfirst}
\addtocontents{toc}{\protect\setcounter{tocdepth}{-1}}

\begin{abstract}
AI dose-prediction models generate patient-specific 3D dose distributions from patient anatomy, providing an important intermediate representation for automated radiotherapy planning. 
However, an AI-predicted dose must undergo downstream optimization to produce a physically constrained treatment plan, a process that requires balancing competing planning objectives.
We present RadOncGym, an agent-based framework that enables an LLM agent to iteratively tune 12 planning-objective weights for dose-mimicking optimization. To reduce iterative evaluation cost, RadOncGym incorporates a surrogate verifier that provides rapid feedback while allowing the agent to decide when to request a full planning solve within a fixed solve budget. We evaluate RadOncGym on 60 OpenKBP head-and-neck patients, each paired with 10 predictions from OpenKBP-Opt dose-prediction sets, yielding 600 patient–prediction contexts per experimental configuration.
Under RadOncGym’s 16-turn setting with a three-solve budget, a GPT-5.6 Sol agent reduces mean clinical-criterion violation by 0.6131 Gy and improves DVH and dose scores by 0.5986 and 0.2204 Gy relative to leaving all 12 weights fixed at 1, outperforming all evaluated non-agent weight-tuning methods under the same solve budget. It remains competitive with the 16-solve variants of these methods while requiring 75.0–77.6\% less normalized runtime.
RadOncGym demonstrates the potential of LLM agents as optimization planners in downstream radiotherapy plan optimization and provides a public testbed for their systematic evaluation.
Code and configurations are available at \href{https://github.com/miaot-usc/RadOncGym}{https://github.com/miaot-usc/RadOncGym}.

\end{abstract}

\section{Introduction}
\label{sec:intro}

\begin{figure}[t]
    \centering
    \includegraphics[width=\columnwidth]{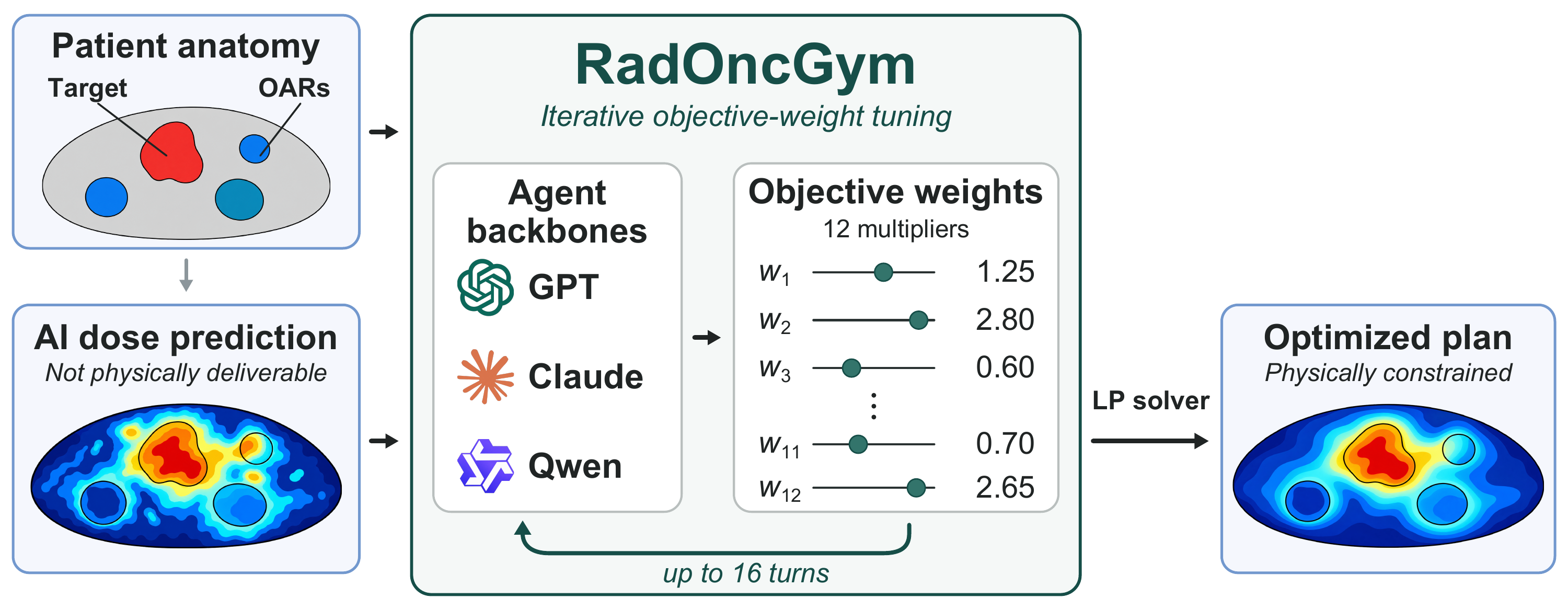}
    \caption{\textbf{Overview of dose-mimicking optimization in RadOncGym.}
    Given patient anatomy and an AI-predicted dose distribution, RadOncGym iteratively adapts 12 planning-objective weights that control how the dose-mimicking optimizer generates a physically constrained treatment plan.}
    \label{fig:problem_overview}
    \vspace{-2mm}
\end{figure}

AI dose-prediction models estimate patient-specific 3D radiotherapy dose distributions from patient anatomy~\cite{babier2021openkbp,gao2023flexible,any2any2026}.
However, dose prediction and treatment-plan optimization are distinct stages of the planning pipeline: an AI-predicted dose distribution must still undergo downstream optimization to produce a physically constrained plan that balances target coverage against sparing of surrounding organs at risk~\cite{babier2022openkbpopt} (Figure~\ref{fig:problem_overview}).
In general, the resulting plan depends on how planning objectives are defined and weighted. Dose-mimicking optimization specializes this stage by using the AI-predicted dose to define patient-specific objective targets.
Within this formulation, a fixed weighting scheme may not be equally suitable across patients or across predictions generated by different dose-prediction models.
We therefore ask: given an AI-predicted dose distribution, how could planning-objective weights be adapted to improve the resulting plan?

Prior work has explored automated planning-parameter tuning using black-box optimization, including random and Bayesian search, as well as reinforcement learning and LLM-based planning agents~\cite{maass2022randomsearch,huang2022metaplanner,wang2023radiotherapybo,shen2019intelligent,shen2020vtpn,abrar2025acer,liu2025gptradplan,gptplan2025,dola2025,zeroshotrtagents2025,sage2025,wei2025cervical,m2m2026}.
Although these methods differ in how they propose candidate planning parameters, evaluating a candidate typically requires running the underlying treatment-planning optimizer and evaluating the resulting plan~\cite{maass2022randomsearch,wang2023radiotherapybo,liu2025gptradplan,dola2025}.
Repeated high-fidelity planning solves can therefore dominate the cost of the search, creating a trade-off between the number of candidate configurations explored and the fidelity of their evaluation.
This motivates a multi-fidelity strategy that evaluates candidate configurations rapidly while reserving expensive planning solves for a selected subset.

We introduce RadOncGym, an agent-based framework and benchmark for dose-mimicking optimization.
Given a patient--prediction context, an LLM agent iteratively proposes a 12-dimensional vector of planning-objective weights. Every proposal receives rapid surrogate feedback, while the agent jointly decides how to update the weights, when to spend its remaining solve budget on a full Gurobi solve, and when to terminate the search. This multi-fidelity interaction enables the agent to explore candidate weight vectors using surrogate feedback while reserving full planning solves for candidates it considers informative or promising.

Across 600 patient--prediction contexts, the GPT-5.6 Sol configuration reduces mean clinical-criterion violation by 0.6131 Gy under a three-solve budget, outperforming all evaluated non-agent weight-tuning methods under the same solve budget.

\noindent\textbf{Our contributions are threefold:}
\begin{itemize}
    \item We introduce RadOncGym, a public agent environment and standardized benchmark for dose-mimicking optimization, enabling controlled comparison of agent configurations and non-agent weight-tuning methods.

    \item We develop a budget-aware agent loop that combines rapid surrogate feedback with selectively requested full planning solves, allowing the agent to allocate a fixed solve budget across iterations.

    \item We demonstrate that LLM agents within RadOncGym's budget-aware optimization loop can achieve a favorable performance--cost trade-off, outperforming matched-budget non-agent methods and remaining competitive with their higher-budget variants while requiring up to 77.6\% less normalized runtime.
\end{itemize}

\section{Related Work}
\label{sec:related}

\subsection{Dose Prediction}
Deep learning has substantially advanced voxel-wise dose prediction. CNN-based models~\cite{kearney2018dosenet,nguyen2019feasibility,yan2021utilizing} have been applied across disease sites, including prostate~\cite{nguyen2020incorporating}, lung~\cite{barragan2019three,gao2023flexible}, esophageal~\cite{zhang2020predicting}, and head-and-neck~\cite{nguyen20193d, liu2021cascade, gheshlaghi2024cascade}. Generative methods based on GAN~\cite{kearney2020dosegan} and diffusion models~\cite{zhang2024dosediff,feng2023diffdp,fu2024md} have also been explored.

\subsection{Automated Radiotherapy Plan Optimization}
A predicted dose distribution, however, is not necessarily achievable under a fixed beam configuration and physical planning constraints. Dose-mimicking optimization uses the predicted distribution as a reference and optimizes a physically constrained plan that approximates it while balancing clinical objectives. OpenKBP-Opt~\cite{babier2022openkbpopt} provides a public formulation of this downstream stage, in which each patient--prediction pair defines a separate planning context. RadOncGym operates at this downstream stage, adapting objective weights without modifying the prediction model, and is therefore orthogonal to dose-prediction methods.

Prior work has optimized planning parameters using black-box search and Bayesian optimization, including Gaussian-process and high-dimensional methods~\cite{maass2022randomsearch,wang2023radiotherapybo,eriksson2021saasbo}.
Other approaches use reinforcement-learning planners~\cite{shen2019intelligent,shen2020vtpn,abrar2025acer} or LLM-based agents~\cite{m2m2026,liu2025gptradplan,sage2025,dola2025,gptplan2025,wei2025cervical,zeroshotrtagents2025} to update planning parameters sequentially.
Despite their different proposal mechanisms, these methods generally require a full solve to evaluate each candidate, making iterative search costly. Many prior studies have also been evaluated on institution-specific cohorts~\cite{wang2023radiotherapybo,liu2025gptradplan,dola2025}, limiting direct comparison across methods.

Building on OpenKBP-Opt~\cite{babier2022openkbpopt}, RadOncGym compares agent and non-agent weight-tuning methods under shared patient--prediction contexts, objective-weight space, planning backend, and solve budget, while using surrogate feedback to reduce reliance on full solves.

\section{Methods}
\label{sec:methods}

\begin{figure*}[!tbp]
    \centering
    \includegraphics[width=0.98\textwidth]{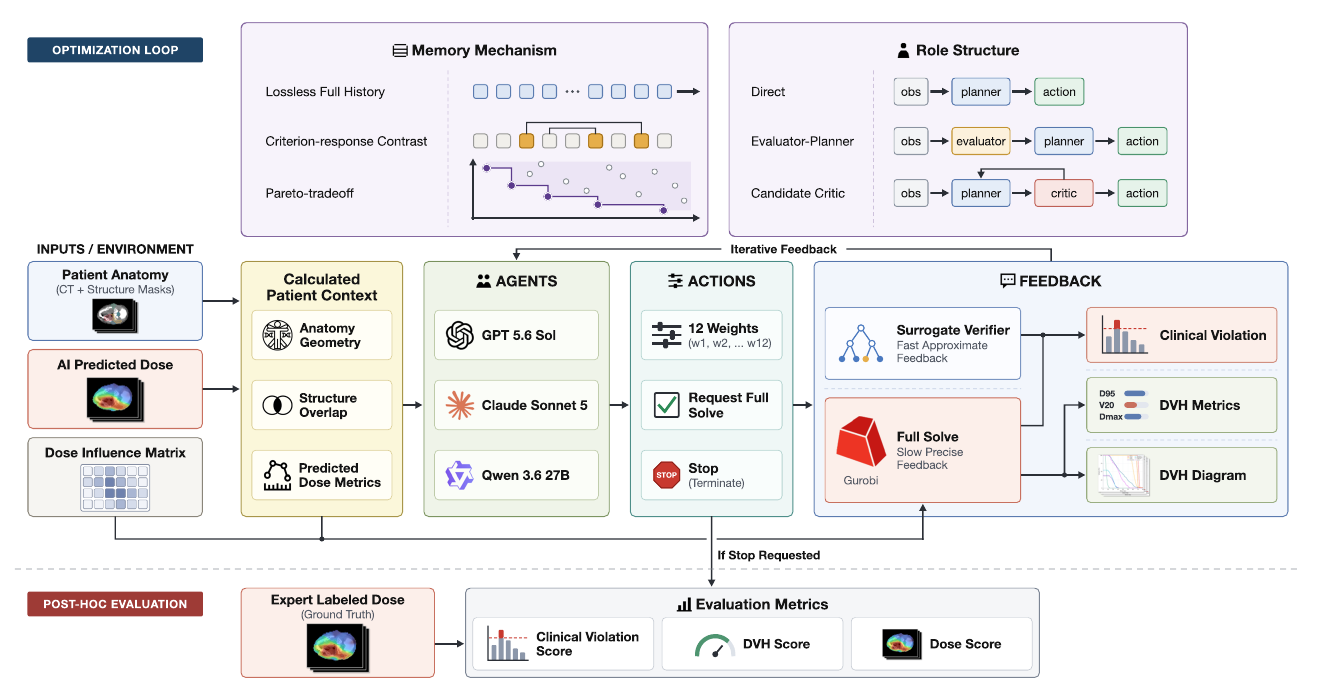}
    \caption{\textbf{RadOncGym environment and agent optimization loop.}
    Context-specific patient anatomy and AI-predicted dose inputs are loaded once.
    The agent iteratively proposes objective weights, may request a full solve, and may stop optimization.
    Every proposal receives feedback from a surrogate verifier, while full solves are requested selectively under a limited budget. 
    DVH and dose scores are computed after plan selection.}
    \label{fig:radoncgym_overview}
\end{figure*}

\subsection{Problem Formulation}
\label{sec:problem_formulation}
Within each patient--prediction context provided by OpenKBP-Opt, let \(a\) denote patient anatomy, \(A_a\) the patient-specific dose-influence matrix, and \(\hat{d}\) an AI-predicted 3D dose distribution. OpenKBP-Opt uses a fixed geometry of nine equally spaced coplanar 6~MV beams, each discretized into beamlets~\cite{babier2022openkbpopt}. Let \(\mathbf{x}\) denote the corresponding vector of beamlet intensities. The resulting dose distribution is
\begin{equation}
    d(\mathbf{x}) = A_a\mathbf{x}.
\end{equation}

A full solve for a planning-objective weight vector \(\mathbf{w}\) computes
\begin{equation}
    \mathbf{x}^{\star}(\mathbf{w})
    \in
    \arg\min_{\mathbf{x} \in \mathcal{X}(a)}
    \mathcal{L}_{\mathrm{DM}}
    \left(d(\mathbf{x}), \hat{d}; \mathbf{w}\right),
\end{equation}
where \(\mathcal{X}(a)\) denotes the feasible set of beamlet intensities under the OpenKBP-Opt planning constraints and \(\mathcal{L}_{\mathrm{DM}}\) denotes the weighted dose-mimicking loss. We denote the resulting optimized dose distribution by \(d_{\mathbf{w}}(a,\hat{d})\).

The vector \(\mathbf{w}\) is the planning parameter exposed to the outer RadOncGym loop. Its 12 entries are planning-objective weights associated with coverage of the three planning target volumes (PTVs), sparing of seven organs at risk (OARs), ring falloff, and posterior-neck sparing. 

\subsection{RadOncGym Environment}
\label{sec:environment_overview}

As shown in Figure~\ref{fig:radoncgym_overview}, RadOncGym comprises an agent-facing environment, a surrogate verifier, and a Gurobi planning backend.  The environment derives structured planning summaries from patient anatomy and AI-predicted dose to form the agent's initial observation. 

At each turn, the agent either stops optimization or proposes 12 planning-objective weights \(\mathbf{w}\in[0.25,4.0]^{12}\) and decides whether to request a full solve. Every proposal receives a surrogate estimate of clinical violation (described in Section~\ref{sec:clinical_metric}) and its uncertainty. If a full solve is requested and budget remains, the environment additionally runs Gurobi to generate an optimized dose plan and returns exact clinical-criterion violations, plan-specific DVH statistics, and a rendered DVH diagram calculated from the optimized plan. The next observation contains the planning summaries, interaction history, returned feedback, and remaining solve budget. This loop continues until the agent stops or the turn horizon is reached. The environment then selects the candidate with the lowest observed mean clinical-criterion violation among those successfully evaluated by a full solve. The selected plan is evaluated using the benchmark endpoints, with expert-reference information used only for post-hoc DVH and dose scoring and never exposed during optimization. 

We implement three memory mechanisms for representing prior optimization turns. Lossless full-history memory retains all previous weight proposals and feedback in chronological order. Criterion-response contrast memory~\cite{dola2025} retains recent turns and the current best candidates, together with contrasting best- and worst-response examples for the dominant clinical-criterion violation. Pareto-tradeoff memory~\cite{knowles1999paes} maintains a Pareto archive of full-solve candidates based on their mean and individual clinical-criterion violations.

We implement three role structures for organizing model calls. Direct planner uses one planner call per environment turn. Evaluator planner~\cite{liu2025gptradplan} first produces a structured assessment of the current observation and then provides it to the planner. Candidate critic follows the self-critique approach of CRITIC~\cite{gou2024critic} and reviews only proposals that request a full solve. The critic may accept the proposal or request a revision before it is executed.

\subsection{Surrogate Verifier}
\label{sec:clinical_metric}

The clinical criteria are not constraints in the dose-mimicking objective above and are scored only after the solve, so a weight vector's clinical quality is observable only after a full Gurobi solve, which dominates the cost of the loop. RadOncGym provides a surrogate verifier that scores proposals between full solves, letting a search method rank candidates and spend its budget on the most promising. Here, “verifier” denotes an approximate plan-quality scorer rather than a physical-feasibility checker.

The clinical objective is the mean positive violation of the 10 OpenKBP-Opt clinical criteria in Table~\ref{tab:clinical_criteria}. For criterion $c$ with measured metric $m_c(d)$, threshold $b_c$, and orientation $s_c\in\{+1,-1\}$, over the applicable set $\mathcal{C}(a)$,

\begin{equation}
J_{\mathrm{clin}}(d)
=
\frac{1}{|\mathcal{C}(a)|}
\sum_{c\in\mathcal{C}(a)}
\max\!\left(
s_c\left(m_c(d)-b_c\right),\,0
\right),
\label{eq:clinical_violation}
\end{equation}

All quantities are in Gy, and $J_{\mathrm{clin}}=0$ means every applicable criterion is satisfied; criteria for structures unavailable in a patient are omitted from $\mathcal{C}(a)$.

Results are reported relative to the \emph{all-ones initialization} $d_{\mathbf{1}}$, the context-matched plan with all 12 weights equal to 1 and the neutral initialization of OpenKBP-Opt~\cite{babier2022openkbpopt}: $\Delta J_{\mathrm{clin}}(d) = J_{\mathrm{clin}}(d_{\mathbf{1}}) - J_{\mathrm{clin}}(d)$, so positive values indicate improvement. The standardized OpenKBP DVH and dose scores~\cite{babier2021openkbp,babier2022openkbpopt} follow the same convention and are computed only after plan selection.

\begin{table}[!htbp]
    \centering
    \small
    \setlength{\tabcolsep}{5pt}
    \renewcommand{\arraystretch}{1.05}
    \begin{tabular}{@{}lc@{\hskip 8pt}|@{\hskip 8pt}lc@{}}
        \toprule
        \textbf{Structure} & \textbf{Criterion}
        & \textbf{Structure} & \textbf{Criterion} \\
        \midrule
        Brainstem     & $D_{0.1\mathrm{cc}}\leq 50$
        & Mandible      & $D_{0.1\mathrm{cc}}\leq 73.5$ \\
        Spinal cord   & $D_{0.1\mathrm{cc}}\leq 45$
        & Larynx        & $D_{\mathrm{mean}}\leq 45$ \\
        Right parotid & $D_{\mathrm{mean}}\leq 26$
        & PTV56         & $D_{99}\geq 53.2$ \\
        Left parotid  & $D_{\mathrm{mean}}\leq 26$
        & PTV63         & $D_{99}\geq 59.9$ \\
        Esophagus     & $D_{\mathrm{mean}}\leq 45$
        & PTV70         & $D_{99}\geq 66.5$ \\
        \bottomrule
    \end{tabular}
    \caption{\textbf{Clinical criteria used to compute mean clinical-criterion violation.}
    Thresholds in Gy, following OpenKBP-Opt~\cite{babier2022openkbpopt}; upper-bound criteria take $s_c=+1$ in Eq.~\eqref{eq:clinical_violation} and the three $D_{99}$ criteria $s_c=-1$.}
    \label{tab:clinical_criteria}
\end{table}

The surrogate verifier is an ensemble of 24 histogram-based gradient-boosting regressors over 309 features, all available without a solve: the 12 proposed weights together with region-of-interest geometry, CT summaries, and AI-predicted-dose summaries, the complete list is in the supplementary material. We report a scalar approximating $\Delta J_{\mathrm{clin}}(d_{\mathbf{w}})$ together with an uncertainty estimate, so positive predictions indicate expected improvement over the all-ones initialization, and verifier error is directly comparable to the improvements of Section~\ref{sec:experiments}. It returns no dose, DVH, or per-criterion prediction.

Training targets come from a fixed offline design, not from agent trajectories. Each patient receives the all-ones initialization and 15 vectors from an independently scrambled Sobol design in log-weight space, $w_j=0.25\cdot16^{u_j}$ with $u_j\in[0,1]$, giving log-uniform marginals on $[0.25,4.0]$; these 16 vectors are reused across the patient's ten prediction contexts. Each is solved with Gurobi and scored by Eq.~\eqref{eq:clinical_violation}, yielding 4,618 and 1,440 usable rows from 4,800 and 1,600 attempts over 30 training and 10 validation patients. The separate 60-patient benchmark cohort is patient-disjoint from both. The ensemble is trained with scikit-learn and combines six gradient-boosting configurations at four seeds each, selected on the validation patients, and is frozen throughout: at inference it does not sample but scores whatever vector the agent or optimizer proposes.

\section{Experiments and Results}
\label{sec:experiments}

\subsection{Experimental Setup}

We evaluate RadOncGym on 60 OpenKBP evaluation patients using the top 10 AI dose-prediction sets from OpenKBP-Opt, yielding 600 patient--prediction contexts per complete configuration~\cite{babier2021openkbp,babier2022openkbpopt}. (Please see the supplementary material for a comparison of baseline plan quality across sets.)

The primary evaluation uses a 16-turn horizon and a budget of at most three full solves per context. A context-matched all-ones plan is precomputed as a common reference and fallback. This reference solve is excluded from the three-solve budget, and its runtime is reported separately in Figure~\ref{fig:cost_performance}. For each method, the reported plan is the successfully full-solved candidate with the lowest observed \(J_{\mathrm{clin}}\); if no full solve succeeds, the all-ones plan is used. DVH and dose scores are computed only after plan selection and never influence optimization or selection.

We use Sobol search as the matched baseline and CMA-ES, Exact-GP LogEI, and SAAS-BO as black-box comparators, collectively referred to as non-agent methods. All methods share the same patient–prediction contexts, objective-weight space, surrogate verifier, Gurobi backend, turn horizon, and solve budget. Agents choose which proposals receive full solves, whereas the standard non-agent methods use surrogate feedback for the first 13 turns and spend their three full solves on terminal candidate verification. Non-agent methods can only receive the scalar clinical-criterion objective; agents additionally receive per-criterion and DVH feedback after successful full solves.

Table~\ref{tab:controls} includes three controls targeting differences in solve timing, feedback content, and horizon length. Adaptive LogEI augments Exact-GP LogEI with a deterministic fidelity-selection controller that interleaves full solves using an online bias-corrected estimate of the full-solve objective. Fixed clinical-only GPT-5.6 Sol uses the same 13-surrogate-then-3-solve schedule and scalar feedback as the non-agent methods, without adaptive solve timing or per-criterion and DVH feedback. Extended Exact-GP LogEI increases the horizon to 64 turns, comprising 61 surrogate evaluations followed by the same three full solves. The complete controller and scheduling rules are provided in the supplementary material.

We additionally evaluate the non-agent methods and a representative GPT-5.6 Sol agent with up to 16 full solves under the same 16-turn horizon.

The memory–role comparison evaluates GPT-5.6 Sol across the full Cartesian product of three memory mechanisms and three role structures. For the cross-model comparison, we fix the agent configuration to criterion-response contrast memory with the candidate-critic role and vary the underlying language model among GPT-5.6 Sol, Claude Sonnet 5, and Qwen3.6-27B, each using its provider-default reasoning setting. 
Outcomes are evaluated using improvements in clinical violation, DVH score and dose score, denoted $\Delta$Clin, $\Delta$DVH, and $\Delta$Dose, respectively.
Min/ctx is normalized runtime per patient--prediction context. We compute uncertainty intervals using patient ID as the resampling unit, treating the 60 patients rather than the 600 contexts as independent.

All methods share the frozen trained surrogate verifier and the evaluation protocol described above; all three LLM backbones support image input and receive the rendered DVH diagram when available. Additional details of LLM usage are provided in the supplementary material.

\subsection{Main Comparison}
\label{sec:main_comparison}

\begin{table}[!tbp]
    \centering
    \scriptsize
    \setlength{\tabcolsep}{1.5pt}
    \renewcommand{\arraystretch}{1.03}

    \begin{tabular}{@{}lcccccc@{}}
        \toprule
        \textbf{Method}
        & \textbf{\shortstack{Tot.\\turns}}
        & \textbf{\shortstack{Full\\solves}}
        & \textbf{$\Delta$Clin $\uparrow$}
        & \textbf{$\Delta$DVH $\uparrow$}
        & \textbf{$\Delta$Dose $\uparrow$}
        & \textbf{\shortstack{Min/\\ctx $\downarrow$}} \\
        \midrule

        Sobol
        & 16 & 3
        & +0.084 & +0.087 & +0.008 & 21.75 \\
        CMA-ES
        & 16 & 3
        & +0.083 & \textbf{+0.259} & +0.037 & 22.39 \\
        Exact-GP LogEI
        & 16 & 3
        & +0.155 & +0.247 & +0.088 & \textbf{20.21} \\
        SAAS-BO
        & 16 & 3
        & \textbf{+0.228} & -0.075 & \textbf{+0.096} & 20.22 \\

        \midrule

        Sobol
        & 16 & 16
        & +0.350 & +0.361 & +0.088 & 110.46 \\
        CMA-ES
        & 16 & 16
        & +0.447 & +0.537 & \textbf{+0.155} & \textbf{109.91} \\
        Exact-GP LogEI
        & 16 & 16
        & +0.577 & +0.536 & +0.127 & 112.06 \\
        SAAS-BO
        & 16 & 16
        & \textbf{+0.649} & \textbf{+0.660} & +0.139 & 122.68 \\

        \midrule

        \textbf{GPT-5.6 Sol}
        & 16 & 3
        & \textbf{+0.613} & \textbf{+0.599} & \textbf{+0.220} & 27.45 \\
        Claude Sonnet 5
        & 16 & 3
        & +0.486 & +0.342 & +0.182 & 31.79 \\
        Qwen3.6-27B
        & 16 & 3
        & +0.405 & +0.175 & +0.151 & \textbf{22.73} \\

        \midrule

        \textbf{GPT-5.6 Sol}
        & 16 & 16
        & \textbf{+0.690} & \textbf{+0.692} & \textbf{+0.223} & 136.33 \\
        \bottomrule
    \end{tabular}

    \caption{\textbf{Main comparison of benchmark improvement and normalized runtime.}
    Blocks separate non-agent and agent families at each budget.
    The 16-solve rows are upper-budget references, not part of the matched three-solve comparison.}
    \label{tab:main_results}
\end{table}
\begin{table}[!tbp]
    \centering
    \scriptsize
    \setlength{\tabcolsep}{1.5pt}
    \renewcommand{\arraystretch}{1.03}

    \begin{tabular}{@{}lcccccc@{}}
        \toprule
        \textbf{Control}
        & \textbf{\shortstack{Tot.\\turns}}
        & \textbf{\shortstack{Full\\solves}}
        & \textbf{$\Delta$Clin $\uparrow$}
        & \textbf{$\Delta$DVH $\uparrow$}
        & \textbf{$\Delta$Dose $\uparrow$}
        & \textbf{\shortstack{Min/\\ctx $\downarrow$}} \\
        \midrule

        Adaptive LogEI
        & 16 & 3
        & +0.236 & +0.070 & +0.039 & 21.65 \\
        GPT-5.6 Sol (fixed, clin.)
        & 16 & 3
        & +0.590 & +0.596 & +0.198 & 21.25 \\
        Exact-GP LogEI (ext.)
        & 64 & 3
        & +0.484 & +0.418 & +0.138 & 37.55 \\
        \bottomrule
    \end{tabular}

    \caption{\textbf{Controls targeting full-solve timing, feedback content, and horizon length.}
    GPT-5.6 Sol (fixed, clin.) uses only scalar clinical feedback, with 13 surrogate turns followed by three full solves; Exact-GP LogEI (ext.) uses 64 total turns.}
    
    \label{tab:controls}
\end{table}

\begin{figure}[!tbp]
    \centering
    \includegraphics[width=\columnwidth]{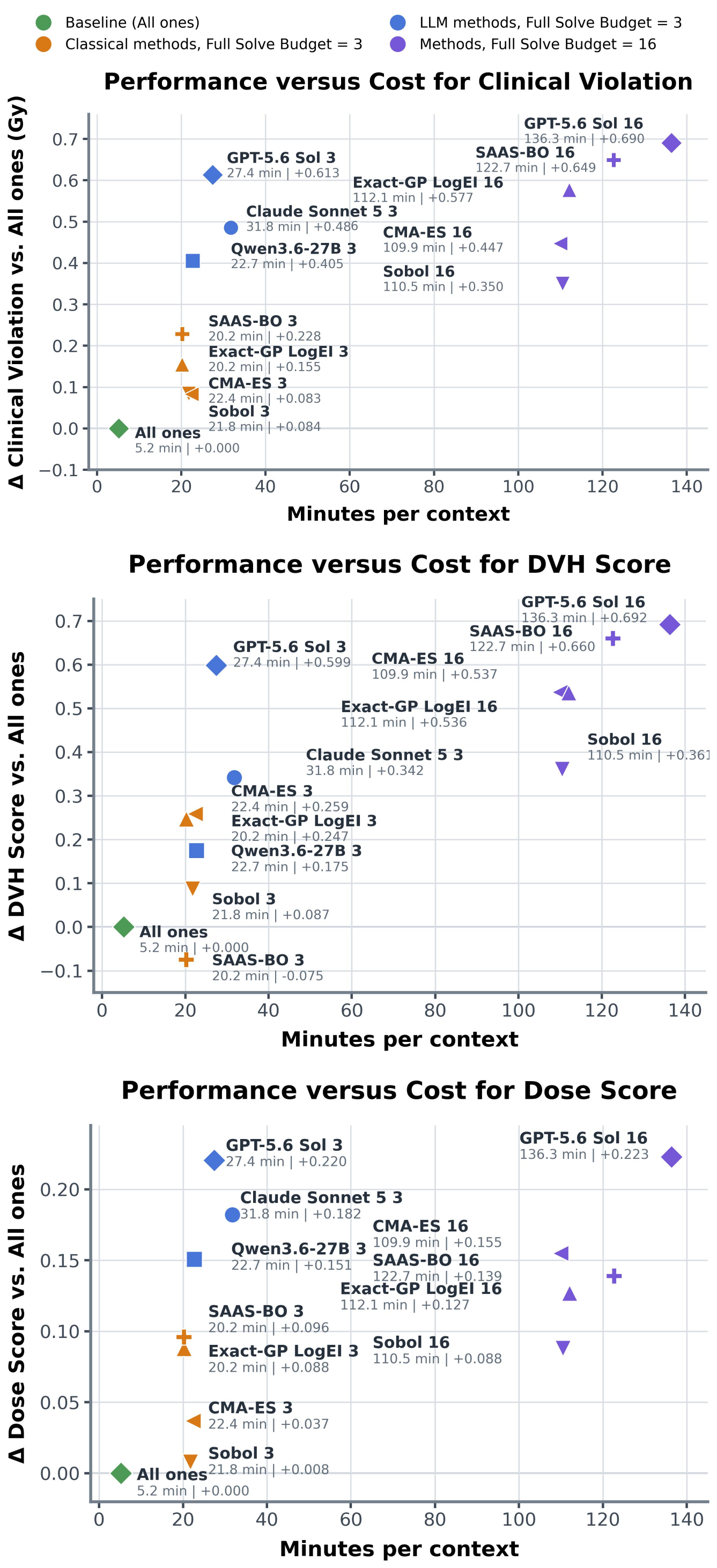}

\caption{\textbf{Cost--performance trade-off.}
Improvement over the context-matched all-ones initialization versus normalized runtime per patient--prediction context, shown for clinical-criterion violation (top), DVH score (middle), and dose score (bottom).
All improvements are in Gy and higher is better on all three endpoints, so points toward the upper left are preferable.
Both the three-solve and 16-solve conditions of Table~\ref{tab:main_results} are included; the all-ones reference solve is excluded from the solve budget and is plotted separately at 5.2 min/context.}
    \label{fig:cost_performance}
\end{figure}

As shown in Table~\ref{tab:main_results} and Figure~\ref{fig:cost_performance}, under the common 16-turn horizon and three-solve budget, GPT-5.6 Sol reduces mean clinical-criterion violation by $0.613$ Gy and improves DVH score and dose score by $0.599$ and $0.220$ Gy, respectively, whereas the best non-agent improvement for each endpoint is $+0.228$, $+0.259$, and $+0.096$ Gy, attained respectively by SAAS-BO, CMA-ES, and SAAS-BO; the adaptive non-agent control, Adaptive LogEI, reaches $+0.236$ Gy on the clinical-criterion endpoint. This comparison matches methods on maximum turn horizon rather than actual surrogate-query count, and the horizon materially affects the result: extending Exact-GP LogEI to 61 surrogate turns before the same three solves raises its clinical-criterion improvement from $+0.155$ to $+0.484$ Gy, narrowing but not closing the gap. Accordingly, the matched-budget comparison is specific to the 16-turn protocol; the extended-horizon control evaluates performance under a larger surrogate-query budget.

For GPT-5.6 Sol with contrast memory and the candidate-critic role, the mean improvements are $+0.613$ Gy for clinical-criterion violation (95\% CI $[0.490,0.729]$), $+0.599$ Gy for DVH score (95\% CI $[0.484,0.718]$), and $+0.220$ Gy for dose score (95\% CI $[0.126,0.322]$).
Paired patient-clustered intervals favor GPT-5.6 Sol over all four non-agent methods at three solves, Adaptive LogEI, and Qwen3.6-27B on all three endpoints; detailed marginal and paired intervals are reported in the supplementary material.

The controls in Table~\ref{tab:controls} clarify the sources of this difference. Adaptive LogEI improves clinical-criterion reduction relative to fixed-schedule Exact-GP LogEI, but produces smaller DVH and dose improvements, indicating that adaptive fidelity selection primarily benefits the scalar objective it observes. Fixed clinical-only GPT-5.6 Sol achieves nearly the same performance as the full agent on all three endpoints ($+0.590$ versus $+0.613$ Gy on clinical-criterion violation, $+0.596$ versus $+0.599$ Gy on DVH score, and $+0.198$ versus $+0.220$ Gy on dose score). These results suggest that iterative agent-based weight selection contributes substantially beyond shared verifier access, whereas adaptive solve timing and richer per-criterion and DVH feedback account for only a small share of the agent's advantage under this protocol.

Under the three-solve budget, GPT-5.6 Sol remains competitive with the 16-solve non-agent methods while requiring 75.0–77.6\% less normalized runtime. When allowed 16 full solves, GPT-5.6 Sol achieves improvements of +0.690, +0.692, and +0.223 Gy on the three endpoints, exceeding all evaluated 16-solve non-agent methods. Model-invocation counts, solver failures, and post-hoc endpoint audits are reported in the supplementary material.

\paragraph{Constraint satisfaction}
To complement mean clinical-criterion violation, we evaluate pooled individual-criterion satisfaction, strict all-criteria feasibility, and the largest remaining criterion violation in each selected plan. For the expert reference, all-ones plans, and GPT-5.6 Sol, respectively, individual-criterion satisfaction is 73.7\%, 77.9\%, and 82.0\%, while strict all-criteria feasibility is 15.0\%, 13.0\%, and 20.5\%. The two endpoints are distinct, and our expert-reference figure differs from the 69.7\% OpenKBP-Opt reports for its full cohort~\cite{babier2022openkbpopt} because we evaluate a 60-patient subset.
Under the three-solve budget, strict all-criteria feasibility is 14.3\% for Sobol, 14.7\% for CMA-ES, 13.7\% for Exact-GP LogEI, 16.2\% for SAAS-BO, 17.5\% for Adaptive LogEI, 18.7\% for Claude Sonnet 5, and 18.5\% for Qwen3.6-27B.
For each selected plan, we identify the largest positive violation among its applicable criteria and then average this worst remaining violation across contexts. This mean decreases from 10.01 Gy for the all-ones initialization to 6.65 Gy for GPT-5.6 Sol. Patient-clustered intervals, full per-method maximum-violation summaries, exact aggregate counts, and per-criterion distributions are reported in the supplementary material.

Figure~\ref{fig:cost_performance} plots improvement against normalized runtime, using the runtimes of Table~\ref{tab:main_results}.

\subsection{Memory and Role-Structure Comparison}
\label{sec:memory_role}

Table~\ref{tab:gpt_memory_role} compares all nine memory--role configurations using GPT-5.6 Sol with its provider-default reasoning setting, evaluated on the same 600 patient--prediction contexts.
\begin{table}[!tbp]
    \centering
    \scriptsize
    \setlength{\tabcolsep}{3pt}
    \renewcommand{\arraystretch}{1.05}

    \begin{tabular}{@{}llccc@{}}
        \toprule
        \textbf{Memory}
        & \textbf{Role}
        & \textbf{$\Delta$Clin $\uparrow$}
        & \textbf{$\Delta$DVH $\uparrow$}
        & \textbf{$\Delta$Dose $\uparrow$} \\
        \midrule

        Lossless full history
        & Direct planner
        & +0.608 & \textbf{+0.612} & +0.226 \\
        & Evaluator planner
        & +0.541 & +0.504 & +0.216 \\
        & Candidate critic
        & +0.603 & +0.595 & \textbf{+0.233} \\

        \midrule

        Criterion-response contrast
        & Direct planner
        & +0.610 & +0.606 & +0.220 \\
        & Evaluator planner
        & +0.533 & +0.494 & +0.210 \\
        & Candidate critic
        & \textbf{+0.613} & +0.599 & +0.220 \\

        \midrule

        Pareto-tradeoff
        & Direct planner
        & +0.592 & +0.585 & +0.215 \\
        & Evaluator planner
        & +0.522 & +0.461 & +0.209 \\
        & Candidate critic
        & +0.585 & +0.572 & +0.223 \\
        \bottomrule
    \end{tabular}

    \caption{\textbf{GPT-5.6 Sol memory and role-structure comparison.}
    Values report improvement over the context-matched all-ones plan in Gy, across the same 60 patients and 600 patient--prediction contexts; higher is better.
    Bold denotes the best value for each endpoint across the nine configurations.}
    \label{tab:gpt_memory_role}
\end{table}

Performance varies only modestly across the nine configurations, and no memory–role combination is best on all three endpoints. The role structures also incur different model-call costs, so the rows represent complete pipeline configurations rather than controlled estimates of an individual role component. Under criterion-response contrast memory, paired critic-versus-direct-planner intervals cross zero on all three endpoints; we therefore do not infer a robust advantage for the critic. Detailed paired estimates are provided in the supplementary material.

\begin{figure*}[!tbp]
    \centering
    \includegraphics[width=\textwidth]{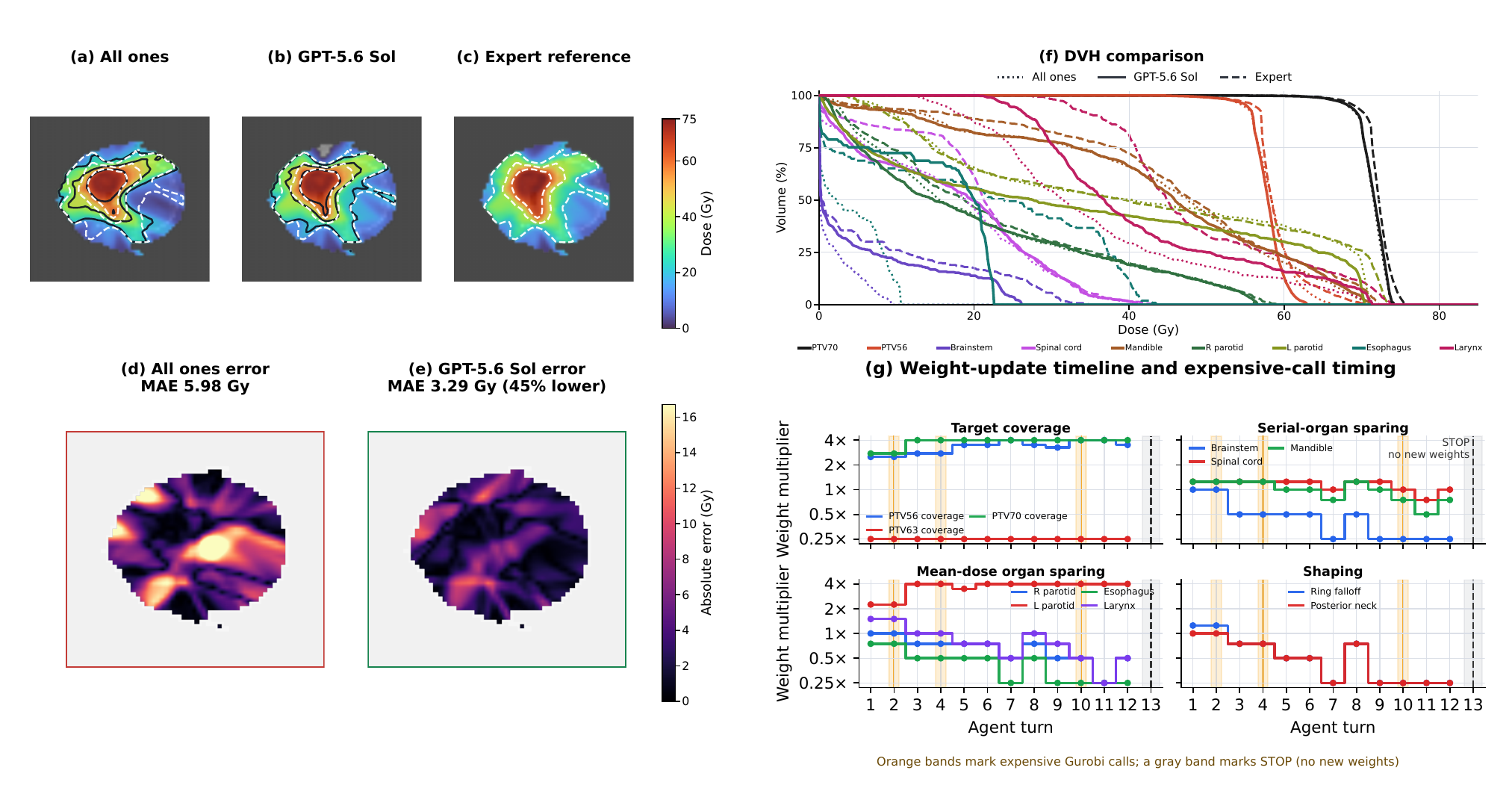}
    \caption{\textbf{Visualization for a representative plan and its weight-update trajectory.}
    (a--c) Dose distributions for the all-ones initialization, the GPT-5.6 Sol optimized plan, and the expert reference.
    Expert-reference isodose contours are shown in white on all three panels; black contours in (a,b) show the generated plans. Closer agreement between black and white contours indicates smaller spatial dose error.
    (d--e) Absolute voxelwise error relative to the expert reference; slice mean absolute error decreases from 5.98 to 3.29 Gy.
    (f) DVH comparison for the three plans.
    (g) Within-context weight history by objective group; vertical bands mark full solves and the final gray marker the stop action.}
    \label{fig:qualitative}
\end{figure*}

\subsection{Surrogate Verifier Utility}
\label{sec:surrogate_verifier}

\begin{figure}[!htbp]
    \centering
    \includegraphics[width=0.95\linewidth]{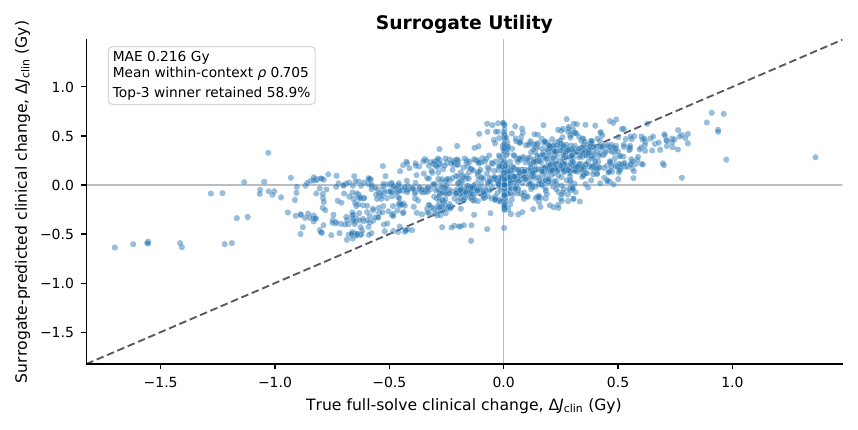}
    \caption{\textbf{Surrogate-verifier utility on validation patients.}
    Surrogate-predicted against true full-solve change in mean clinical-criterion violation relative to the all-ones initialization.}
    \label{fig:surrogate_verifier}
    \vspace{-2mm}
\end{figure}

We evaluate whether the surrogate verifier supports candidate screening between full solves. On validation patients, Figure~\ref{fig:surrogate_verifier} compares the verifier-predicted change in mean clinical-criterion violation with the change obtained from a full solve.
The verifier achieves MAE $=0.216$ Gy and mean within-context Spearman correlation $\rho=0.705$, and the true best candidate falls within its predicted top three in 58.9\% of validation contexts, supporting its use as a low-cost ranking and screening signal.

To test for distribution shift under adaptive optimization, we additionally evaluate the verifier on agent-generated actions that were subsequently sent to Gurobi.
Of 1,765 requested full solves, 1,732 yield paired surrogate and full-solve outcomes after excluding 33 timeouts.
On these adaptive queries the verifier has MAE $=0.358$ Gy and agrees with the full solve on whether a candidate improves over the all-ones initialization in 81.6\% of cases, with the largest turn-binned error during turns 5--8; correlation statistics are reported in the supplementary material.
Predictive error increases on agent-generated queries, but the verifier retains 81.6\% agreement on improvement direction, supporting its use as a screening signal under this distribution shift.

\subsection{Agent Behavior and Weight Adaptation}
\label{sec:loop_behavior}
\label{sec:weight_behavior}

The representative GPT-5.6 Sol configuration typically spends its budget early rather than conserving it. It obtains three successful full solves in 91.8\% of contexts, while 5.8\%, 1.5\%, and 0.8\% use two, one, and zero solves; five contexts have no successful solve and fall back to the all-ones plan.
Explicit stopping occurs in 599 of 600 contexts, with median turns of 2, 4, 7, and 9 for the three full solves and the stop, so the gap before stopping reflects when the agent chooses to end rather than additional optimization.
Carrying the best incumbent forward, mean clinical-criterion improvement reaches 54\% of the final $+0.6131$ Gy after the first full solve and 88\% after the second; among contexts reaching the second and third solves, new candidates replace the incumbent in 77.1\% and 65.5\%, respectively.
Full-solve usage, conditional turn distributions, and stage-wise gains are reported in the supplementary material.
Table~\ref{tab:weight_distribution} summarizes the direction of each final proposal weight relative to the all-ones initialization and its mean value.
The final direction differs substantially across objectives. PTV70 is increased in 96.2\% of contexts, and PTV56 and both parotid weights more often increase than decrease, while the serial-organ, larynx, ring-falloff, and posterior-neck weights more often decrease. PTV63 more often decreases, although its mean final weight stays near one. These patterns describe the joint behavior of the 12-dimensional agent and should not be interpreted as causal effects of individual weights.

\begin{table}[!tbp]
    \centering
    \small
    \setlength{\tabcolsep}{6pt}
    \begin{tabular}{@{}lcc@{}}
        \toprule
        \textbf{Objective} & \textbf{Final $\uparrow$ / $\downarrow$} & \textbf{Mean final weight} \\
        \midrule
        PTV56 coverage        & 69.2\% / 18.8\% & 1.694 \\
        PTV63 coverage        & 32.5\% / 56.5\% & 1.017 \\
        PTV70 coverage        & 96.2\% / 0.8\%  & 2.941 \\
        Brainstem sparing     & 2.3\% / 88.3\%  & 0.472 \\
        Spinal cord sparing   & 11.2\% / 74.2\% & 0.636 \\
        Mandible sparing      & 12.3\% / 74.0\% & 0.621 \\
        Right parotid sparing & 65.7\% / 26.0\% & 2.216 \\
        Left parotid sparing  & 67.3\% / 26.2\% & 2.333 \\
        Esophagus sparing     & 5.2\% / 89.8\%  & 0.411 \\
        Larynx sparing        & 24.5\% / 67.0\% & 0.936 \\
        Ring falloff          & 25.2\% / 65.0\% & 0.769 \\
        Posterior neck sparing & 13.8\% / 74.2\% & 0.623 \\
        \bottomrule
    \end{tabular}
    \caption{\textbf{Planning-objective weight behavior for the representative GPT-5.6 Sol agent.}
    ``Final $\uparrow$ / $\downarrow$'' gives the fraction of the 600 contexts whose final proposal weight ends more than 5\% above or below the all-ones initialization.
    Per-step movement frequencies are in the supplementary material.}
    \label{tab:weight_distribution}
\end{table}

\subsection{Qualitative Plan Visualization}
\label{sec:qualitative}

Figure~\ref{fig:qualitative} presents a representative case in which planning-objective weight tuning brings the optimized dose distribution closer to the expert reference, as reflected by reduced voxelwise error and improved DVH agreement.

\section{Limitations and Future Work}
\label{sec:limitations}

\noindent\textbf{Benchmark scope.}
RadOncGym is evaluated on 60 head-and-neck patients, one fixed OpenKBP-Opt planning formulation, the top 10 OpenKBP-Opt dose-prediction sets, and an action space restricted to 12 planning-objective weights. This fixed scope enables controlled comparison but does not establish generalization to other disease sites, clinical protocols, prediction models, planning formulations, or treatment-planning systems. Future extensions could incorporate standardized adapters for additional public datasets, planning backends, and optimization parameters.

\noindent\textbf{Statistical dependence and run-to-run variability.}
The 600 patient--prediction contexts arise from 60 patients, each paired with 10 dose predictions. We therefore cluster uncertainty estimates by patient to account for the shared anatomy within each group of 10 contexts. All campaigns use one fixed seed, so the reported intervals characterize variation across patients but not variation across repeated runs. Multi-seed experiments would provide a fuller assessment of run-to-run variability.

\noindent\textbf{Comparison interpretation.}
The primary comparison holds the turn horizon, surrogate verifier, and full-solve budget fixed across methods. However, agents receive richer observations and feedback and can choose when to request full solves, whereas the non-agent methods use scalar feedback and a fixed solve schedule. Although the controls in Table~\ref{tab:controls} evaluate some of these differences, the observed performance gap reflects the overall agent pipeline and should not be attributed to the LLM alone.

\noindent\textbf{Clinical interpretation.}
The reported endpoints summarize 10 fixed clinical criteria and agreement with the expert reference dose. The clinical implications of improvements in these benchmark metrics require further study. OpenKBP-Opt produces fluence-based plans under a research dose-influence model and plan-complexity constraint; it does not reproduce the complete clinical workflow of machine-specific sequencing, treatment-planning-system dose recalculation, and quality assurance. The reported improvements should therefore be interpreted as benchmark gains rather than evidence of clinical benefit or deployability. Any generated plan would require standard quality assurance and expert clinical review, and prospective evaluation within clinical planning systems is needed before translation.

\section{Conclusion}
\label{sec:conclusion}

We introduced RadOncGym, an agent-based framework for budget-aware dose-mimicking optimization. By combining surrogate feedback with agent-selected full planning solves, RadOncGym enables an LLM agent to adapt planning-objective weights while allocating a limited solve budget across iterations.

Across 600 patient--prediction contexts, its GPT-5.6 Sol configuration under a three-solve budget improved all three benchmark endpoints, including a 0.6131~Gy reduction in mean clinical-criterion violation. Under the same solve budget, it outperformed all evaluated non-agent weight-tuning methods and remained competitive with their 16-solve variants at 75.0--77.6\% lower normalized runtime. These findings support the use of LLM agents as optimization planners under limited solve budgets. RadOncGym also offers a public testbed for studying agent-based optimization in downstream radiotherapy planning.

\clearpage
\section*{Acknowledgments}
This work was supported by the National Science Foundation (NSF) under grants 2613950 and 2613761.

{\small
\bibliographystyle{ieeenat_fullname}
\bibliography{references}
}

\clearpage
\input{supplementary}
\end{document}